\documentclass[twoside,11pt]{article} 
\pdfoutput=1

\usepackage{jmlr2e}
\usepackage{amsmath}
\usepackage{algorithm,algorithmic}

\newcommand{\qed}{\hfill\BlackBox\\[2mm]}

\ShortHeadings{Reconsidering Dependency Networks}{Takabatake and Akaho}
\firstpageno{1}

\begin{document}

\title{Reconsidering Dependency Networks from an Information Geometry Perspective}

\author{\name Kazuya Takabatake \email k.takabatake@aist.go.jp \\
       \addr Human Informatics and Interaction Research Institute\\
       National Institute of Advanced Industrial Science and Technology\\
       Central 2: 1-1-1 Umezono, Tsukuba, Ibaraki 305-8568, Japan.
       \AND
       \name Shotaro Akaho \email s.akaho@aist.go.jp \\
       \addr Human Informatics and Interaction Research Institute\\
       National Institute of Advanced Industrial Science and Technology\\
       Central 2: 1-1-1 Umezono, Tsukuba, Ibaraki 305-8568, Japan.}
\editor{}

\maketitle

\begin{abstract}
Dependency networks \citep{heckerman2000dependency} are potential probabilistic graphical models for systems comprising a large number of variables.
Like Bayesian networks, the structure of a dependency network is represented by a directed graph, and each node has a conditional probability table.
Learning and inference are realized locally on individual nodes; therefore, computation remains tractable even with a large number of variables.
However, the dependency network's learned distribution is the stationary distribution of a Markov chain called pseudo-Gibbs sampling and has no closed-form expressions.
This technical disadvantage has impeded the development of dependency networks.
In this paper, we consider a certain manifold for each node.
Then, we can interpret pseudo-Gibbs sampling as iterative m-projections onto these manifolds.
This interpretation provides a theoretical bound for the location where the stationary distribution of pseudo-Gibbs sampling exists in distribution space.
Furthermore, this interpretation involves structure and parameter learning algorithms as optimization problems.
In addition, we compare dependency and Bayesian networks experimentally.
The results demonstrate that the dependency network and the Bayesian network have roughly the same performance in terms of the accuracy of their learned distributions.
The results also show that the dependency network can learn much faster than the Bayesian network.
\end{abstract}


\section{Introduction}
The primary purpose of this paper is to investigate dependency networks \citep{heckerman2000dependency} from an information geometry perspective and offers new insights into dependency networks.

Bayesian and Markov probabilistic graphical models \citep[][Chapter 3,4]{koller2009probabilistic} are frequently employed to construct multivariate probabilistic models.
Both Bayesian and Markov graphical models have well-studied theoretical foundations, and their structure/parameter learning and inference algorithms are also well-established.
Here, let $X=\{X_0,...,X_{n-1}\}$ be variables in the entire network and $p^*(X)$ be the true distribution underlying training data.
Then, the basic mechanisms of Bayesian and Markov networks are as follows:
\begin{description}
\item[Learning.] Given training data, the entire network learns $p^*(X)$, that is, the entire network builds a distribution $\pi(X)$ that is an estimation of $p^*(X)$.
\item[Inference.] Given inference condition $V=v$, the entire network computes the posterior distribution $\pi(U|V=v)$, where $U,V$ are certain variables in the network.
\end{description}

In Bayesian network structure learning \citep[][Chapter 17]{koller2009probabilistic}, the network searches the graphical structure $G$ that minimizes $scost(G)$ for a structure cost function $scost$.
The set of possible $G$ becomes extremely large as the number of nodes increases.
Even using greedy algorithms, the network must evaluate vast $scost(G)$ values, which leads to very slow structure learning.

In Markov network parameter learning \citep[][Chapter 20]{koller2009probabilistic}, the network searches parameters $\theta=\{\theta_j\}(j\in J)$ that minimize $pcost(\theta)$ for a parameter cost function $pcost$.
$|J|$(=number of parameters) becomes large as the number of nodes increases.
In such cases, parameter learning becomes a minimization problem in a very high-dimensional space that is intractable even for high-dimensional minimization algorithms such as L-BFGS \citep[][]{liu1989limited}.

Dependency network graphical models \citep{heckerman2000dependency} hold potential for large systems.
Like Bayesian networks, the structure of a dependency network is represented by a directed graph, and each node has a conditional probability table (CPT).
However, differing from Bayesian networks, dependency network graphs can be cyclic.

In dependency network structure learning, the structure cost function $scost$ can be divided into local costs as follows:
$$scost(G)=\sum_i scost_i(Y_i),$$
where $Y_i$ is the local structure of the node $i$ (Section \ref{sec:dependency network structure}).
In this case, we can minimize $scost$ by independently minimizing $scost_i$.
Since the search space for $Y_i$ is much smaller than the search space for $G$,
structure learning remains tractable even in the cases where the number of nodes is large.

In dependency network parameter learning, we can also divide the parameter cost function $pcost$ into local costs as follows:
$$pcost(\theta)=\sum_i pcost_i(\theta_i).$$
Therefore, we can perform parameter learning by independent ``local" learning of $\theta_i$ (Section \ref{sec:Parameter Learning}).
This independence of local learning keeps parameter learning tractable even with a large number of nodes \footnote{Bayesian network parameter learning also has this property.}.

Let $X_{-i}$ be variables in the entire network except for $X_i$, and let $Y_i$ be certain variables included in $X_{-i}$.
For a distribution $p(X)$, $p(X_i|X_{-i})$ is called the {\it full conditional distribution} \citep[][Chapter 5]{gilks1995markov}.
Now, the framework of the dependency networks can be described as follows:
\begin{description}
\item[Learning.] Given training data, each node learns its full conditional distribution independently, that is, each node builds CPT $\theta_i(X_i|Y_i)$ that is an estimation of $p^*(X_i|X_{-i})$.
\item[Inference.] Given CPTs and an inference condition $V=v$, a Markov chain Monte Carlo method called {\it pseudo-Gibbs sampling} \citep{heckerman2000dependency} draws samples from its stationary distribution\footnote{We refer to these samples as {\it output data} (or shortly {\it outputs}) of pseudo-Gibbs sampling.}, and this stationary distribution becomes the estimation of $p^*(U|V=v)$.
Users know the estimation by counting samples in the output data.
\end{description}

Learning the full conditional distribution is a local task of each node; therefore, it is much more tractable than learning the joint distribution.
Various regression techniques can be applied to learn full-conditional distributions.
If a user use a regression technique with input variable selection, the user can ignore most variables in $X_{-i}$ and can approximate $p^*(X_i|X_{-i})$ by $\theta_i(X_i|Y_i)$, where $Y_i$ consists of only several variables.

The inference task is also tractable even in cases where the number of nodes is large because pseudo-Gibbs sampling consists of simple local tasks (Section \ref{sec:pseudo}).

Although the dependency network mechanism consists of such simple local tasks, the distribution represented by the dependency network is complicated.
Let $\pi(X)$ be the joint distribution represented by a graphical model.
For Bayesian networks, $\pi(X)$ is expressed as follows:
\begin{equation}\label{eq:BN represent}
\pi(X)=\prod_i \pi(X_i|Y_i),
\end{equation}
where $Y_i$ denotes the {\it parents} of $X_i$ \citep[][Chapter 2]{koller2009probabilistic}.
For Markov networks, $\pi(X)$ is expressed as follows:
\begin{equation}\label{eq:MN represent}
\pi(X)=\frac1Z\prod_c\phi_c(X_c),
\end{equation}
where $c$ denotes {\it clique} in the network, $X_c$ denotes variables included in $c$, $\phi_c$ denotes {\it clique potential}, and $Z$ denotes {\it partition function} \citep[][Chapter 4]{koller2009probabilistic}.
In contrast, the joint distribution represented by a dependency network is the stationary distribution of a Markov chain called {\it pseudo-Gibbs sampling} \citep{heckerman2000dependency}, and this stationary distribution has no closed-form expressions such as Eqs.\eqref{eq:BN represent} and \eqref{eq:MN represent}.
This disadvantage is significant; for example, the conventional maximum likelihood estimation method is not applicable because the likelihood itself is unknown.
The progress of dependency networks has been disturbed by this problematic property.

\citet{heckerman2000dependency} investigated a specific class of dependency networks, called {\it consistent dependency networks}, which are equivalent to Markov networks, and explained general dependency networks by perturbation from consistent dependency networks.
They showed a theoretical bound for the difference between consistent dependency networks and general dependency networks.
However, they also found that their theoretical bound was too loose in practice.

\citet{takabatake2012constraint} independently studied graphical models, which they referred to as {\it firing process networks}, which are equivalent to general dependency networks.
They considered a certain manifold for each node and interpreted pseudo-Gibbs sampling as iterative m-projections onto these manifolds.
This interpretation offers the following consequences:
\begin{itemize}
\item Providing a theoretical bound for the location where the stationary distribution exists in distribution space;
\item Providing how to construct structure/parameter learning algorithms as optimization problems.
\end{itemize}

The remainder of this paper is organized as follows.
In Section 2, we review the structure and mechanism of dependency networks.
In Section 3, from a perspective of information geometry, we discuss pseudo-Gibbs sampling, which is a Markov chain Monte Carlo method to synthesize a joint distribution from CPTs owned by the nodes.
In Section 4, we derive structure and parameter learning algorithms based on the interpretation obtained in Section 2.
Probabilistic inference in dependency networks is discussed in Section 5.
In Section 6, we generalize the input variables referred by CPT to enhance the expressive power of dependency networks.
In Section 7, we compare dependency and Bayesian networks and examines the behavior of dependency network learning experimentally.
Conclusions and suggestions for future work are provided in Section 8.
Appendix section provides proofs of theorems.

\section{Dependency Networks}\label{sec:dependency}
We explain some notations used in this paper.
A random variable is denoted by a capital letter, such as $X$, and a value that $X$ takes is denoted by a lower case letter, such as $x$.
Concatenated variables such as $(X,Y,Z)$ are shortened and denoted $XYZ$.
Note that concatenated variables are often treated as a single variable.
When $X$ is a concatenation of certain variables, the set of the variables included in $X$ are also denoted by $X$, for example, if $X=X_0X_1X_2X_3$ and $Y=X_1X_2$ then $Y\subseteq X$ and $X\setminus Y=X_0X_3$.
$\left<f(X)\right>_{p(X)}$ denotes the expectation of $f(X)$ with distribution $p(X)$, that is, $\left<f(X)\right>_{p(X)}=\sum_x p(x)f(x)$.
For a distribution $p$, an empirical distribution of data drawn from $p$ is denoted as $\tilde p$.

\subsection{Dependency Network Structure}\label{sec:dependency network structure}
\begin{figure}[H]
\centering\includegraphics{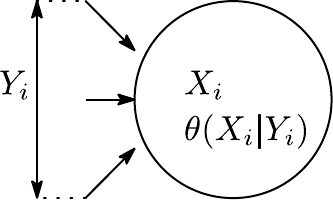}
\caption{Node $i$}\label{fig:node}
\end{figure}
The structure of dependency networks resembles the structure of Bayesian networks.
A dependency network is represented by a directed graph with $n$ nodes indexed by the numbers $0,...,n-1$.
As shown in Figure \ref{fig:node}, each node has a variable $X_i$ and a conditional probability table (CPT) $\theta_i(X_i|Y_i)$, where $Y_i$ denotes certain variables in $X_{-i}$.
We call $Y_i$ {\it input variables} (or simply {\it inputs}) of node $i$.

It is possible to consider that $X_i$ takes continuous values.
In this case, $\theta_i$ is a density function, and CPT is implemented using a regression technique for continuous variables.
However, for simplicity, we focus on the cases where $X_i$ takes discrete finite values in this paper.

The dependency network graph consists of edges directed from node $i$'s input variables to node $i$.
Unlike the Bayesian network, the graph of a dependency network can be cyclic.

\subsection{Pseudo-Gibbs Sampling}\label{sec:pseudo}
Here, we explain how to synthesize a joint distribution from CPTs owned by the nodes.

{\it Pseudo-Gibbs sampling} \citep{heckerman2000dependency} is a Markov chain Monte Carlo method that is applied to synthesize a joint distribution $\pi(X)$ from CPTs owned by the nodes.
The following algorithm is an example of pseudo-Gibbs sampling.
\begin{algorithm}[H]
\caption{Pseudo-Gibbs sampling}
$X=X_0...X_{n-1}$: variables in the entire network\\
$o$: output data\\
$N$: number of samples to be drawn
\begin{algorithmic}[1]
\STATE $o\leftarrow\emptyset$, $X\leftarrow x^0$ \COMMENT{arbitrary initial value}
\FOR{$t=0$ \TO $N-1$}
\STATE Append $X$ into $o$
\STATE Select a node $i$ \label{line:select a node}
\STATE Referring to inputs $Y_i$, draw a sample $x_i$ from distribution $\theta_i(X_i|y_i)$ and let $X_i\leftarrow x_i$ \label{line:fire a node}
\ENDFOR
\RETURN $o$
\end{algorithmic}
\end{algorithm}
There are two ways to select a node in line \ref{line:select a node}.
One is to select a node in a sequential and cyclic manner such as $0,1,...,n-1,0,1,...$.
We refer to this type of pseudo-Gibbs sampling as {\it ordered-pseudo-Gibbs sampling}.
The second option is to randomly select a node $i$ with probability $c_i$ such that
\begin{equation}\label{eq:c_i}
\sum_i c_i=1.
\end{equation}
Typically, $c_i=1/n$ is used unless there is a particular reason (Section \ref{sec:inference}).
We refer to this type of pseudo-Gibbs sampling as {\it random-pseudo-Gibbs sampling}.
The operation in line \ref{line:fire a node} is mentioned frequently in this paper; therefore, we will refer to this operation as {\it firing node $i$}.

Let $O^N=\{X^0,...,X^{N-1}\}$ be $N$ samples of output data---as a sequence of random variables---of pseudo-Gibbs sampling algorithm.
Here, we assume the ergodicity of random-pseudo-Gibbs sampling.
Then, the following consequences hold \citep[][Chapter 3]{gilks1995markov}.
\begin{itemize}
\item There uniquely exists a distribution $\pi(X)$, called {\it stationary distribution}, such that for any initial value $x^0$,
\begin{equation}\label{eq:stationary}
\pi(X)=\lim_{t\to\infty}p(X^t|x^0).
\end{equation}
\item For any real valued function $f(X)$ that satisfies $\left<|f(X)|\right>_{\pi(X)}<\infty$,
\begin{equation}\label{eq:ergodicity}
\lim_{N\to\infty}\frac{f(X^0)+...+f(X^{N-1})}N=\left<f(X)\right>_{\pi(X)}\quad(a.s.).
\end{equation}
\end{itemize}
Substituting $1(X=x)$ (indicator function) for $f$ in Eq.\eqref{eq:ergodicity}, we obtain
\begin{equation}\label{eq:represented}
\lim_{N\to\infty}\frac{N_x}N=\pi(x)\quad(a.s.),
\end{equation}
where $N_x$ denotes the number of occurrences of $X^t=x(t=0,...,N-1)$ in the output dataset.
Equation \eqref{eq:represented} implies that, even if it is not possible to know $\pi(x)$ directly, we can estimate $\pi(x)$ by a Monte Carlo method.

Ordered-pseudo-Gibbs sampling is an {\it inhomogeneous} Markov chain \citep[][Chapter 4]{seneta2006non} and does not have a stationary distribution.
Even in this case, the subsequence $X^iX^{i+n}X^{i+2n}...$ becomes a {\it homogeneous} Markov chain and has a stationary distribution $\pi_i(X)$ under the assumption of ergodicity.
Therefore, the limit in the left side of Eq.\eqref{eq:represented} still exists and
\begin{equation}\label{eq:ordered}
\lim_{N\to\infty}\frac{N_x}N=\frac{\pi_0(x)+...+\pi_{n-1}(x)}n.
\end{equation}

Contrary to pseudo-Gibbs sampling, we refer to usual Gibbs sampling as {\it actual-Gibbs sampling} in this paper.
Actual-Gibbs sampling is a special case of the pseudo-Gibbs sampling such that there exists a joint distribution $p(X)$ that satisfies
\begin{equation}\label{eq:Gibbs}
\theta_i(X_i|Y_i)=p(X_i|X_{-i})
\end{equation}
for all $i$.
In this case, $p$ becomes the stationary distribution because firing any node $i$ does not move $p$.

Here, we build the following rough hypothesis.
\begin{description}
\item[Hypothesis 1.] If every CPT is a good estimation of $p^*(X_i|X_{-i})$, then the stationary distribution of pseudo-Gibbs sampling becomes a good estimation of $p^*(X)$.
\end{description}
Hypothesis 1 plays the central role in dependency network learning.
We will justify this hypothesis in Section \ref{sec:information geometry}.

\subsection{Local Learning}
The learning task involves building a distribution $\pi(X)$ that is a good estimation of the true distribution $p^*(X)$.
In dependency networks, controlling the stationary distribution $\pi(X)$ by adjusting CPTs is difficult.
Therefore, rather than learning $p^*(X)$ directly, each node in the dependency network learns the full-conditional distribution $p^*(X_i|X_{-i})$ independently from training data.
A favorable property here is that node $i$ can learn $p^*(X_i|X_{-i})$ without knowing other nodes' behavior.

Local learning is realized using regression algorithms with input variable selection.
For example, \cite{heckerman2000dependency} used a decision tree algorithm.
In the process of constructing the tree, variables $Y_i$ are automatically selected from $X_{-i}$.

We will present our proposed learning algorithm in Section \ref{sec:learning}.

\subsection{Probabilistic inference by Pseudo-Gibbs sampling}\label{sec:inference}
Let $U$ and $V$ be certain variables in the network.
Probabilistic inference involves estimating the posterior distribution $p^*(U|v)$.
Estimating $p^*(X)$ described in Section \ref{sec:pseudo} is a special case where $U=X,V=\emptyset$.

We can use pseudo-Gibbs sampling again to estimate $p^*(U|v)$ here.
After learning CPTs, the value of $V$ is clamped to $v$, and only the nodes out of $V$ are fired in pseudo-Gibbs sampling.
We refer to this type of pseudo-Gibbs sampling as {\it clamped-pseudo-Gibbs sampling}\footnote{In contrast, we refer to the pseudo-Gibbs sampling described in Section \ref{sec:pseudo} as {\it free-pseudo-Gibbs sampling}.}.
Then, the stationary distribution of clamped-pseudo-Gibb sampling is used as the estimation of $p^*(U|v)$.

The following algorithm is an example of clamped-pseudo-Gibbs sampling.
\begin{algorithm}[H]
\caption{Clamped-pseudo-Gibbs sampling}\label{algo:clamped}
$X=X_0...X_{n-1}$: variables in the entire network\\
$V$: variables to be clamped to $v$\\
$\overline V$: all variables out of $V$\\
$o$: output dataset\\
$N$: number of samples to be drawn
\begin{algorithmic}[1]
\STATE $o\leftarrow\emptyset$, $\overline V\leftarrow \overline v^0$ \COMMENT{arbitrary initial value}, $V\leftarrow v$
\FOR{$t=0$ \TO $N-1$}
\STATE Append $X$ into $o$
\STATE Select a node $i\not\in V$ \label{line:select a node(clamped)}
\STATE Fire node $i$
\ENDFOR
\RETURN $o$
\end{algorithmic}
\end{algorithm}
In line \ref{line:select a node(clamped)}, if we select a node in a random manner (random-pseudo-Gibbs sampling) then the probability to select node $i$ becomes as follows:
$$c_i=\begin{cases}
\frac1{n-|V|} &i\in V\\
0 &i\not\in V
\end{cases},$$
where $|V|$ is the number of nodes in $V$.

Let define as $\overline V:=X\setminus V$ and $Z_i:=X_{-i}\setminus V$.
Note that $\theta_i(X_i|Y_i)$, which is a function of $X_iY_i$, is a special case of $\theta_i(X_i|X_{-i})$, which is a function of $X$, because $X_iY_i\subseteq X$, that is, we can always rewrite $\theta_i(X_i|Y_i)$ as $\theta_i(X_i|X_{-i})$.
Then, the clamped-pseudo-Gibbs sampling is equivalent to the free-pseudo-Gibbs sampling with CPTs $\{\theta_i(X_i|Z_iv)\}(i\in\overline V)$.
Let $\pi_v(\overline V)$ be	the stationary distribution of clamped-pseudo-Gibbs sampling.
If $\theta_i(X_i|X_{-i})$ is a good estimation of $p^*(X_i|X_{-i})$, then $\theta_i(X_i|Z_iv)$ is also considered a good estimation of $p^*(X_i|Z_iv)$.
Therefore, $\pi_v(\overline V)$ is considered  a good estimation of $p^*(\overline  V|v)$ by Hypothesis 1.
Let $N_u$ be the number of samples such that $U=u$ in the $N$ samples of output data of clamped-pseudo-Gibbs sampling.
Then, $N_u/N$ becomes a good estimation of $p^*(u|v)$ for sufficiently large $N$.

Output data obtained by Algorithm \ref{algo:clamped} are far from i.i.d. data.
To make output data closer to i.i.d. data, we use {\it burn-in} and {\it thinning} techniques \citep[Chapter 7 and 8]{gilks1995markov}, that is, we discard the first $b$ samples and use only every $k$-th samples.
Then, Algorithm \ref{algo:clamped} is modified as follows:
\begin{algorithm}[H]
\caption{Clamped-pseudo-Gibbs sampling with burn-in and thinning}\label{algo:clamped with burn-in and thinning}
$X=X_0...X_{n-1}$: variables in the entire network\\
$V$: variables to be clamped to $v$\\
$\overline V$: all variables out of $V$\\
$o$: output dataset\\
$N$: number of samples to be drawn\\
$b$: The first $b$ samples are discarded from the output.\\
$k$: Every $k$-th samples are used for outputs.
\begin{algorithmic}[1]
\STATE $o\leftarrow\emptyset$, $X\leftarrow x^0$ \COMMENT{arbitrary initial value}, $V\leftarrow v$
\FOR{$s=1$ \TO $b$}
\STATE Select a node $i\not\in V$
\STATE Fire node $i$
\ENDFOR
\FOR{$t=0$ \TO $N-1$}
\STATE Append $X$ into $o$
\FOR{$s=1$ \TO $k$}
\STATE Select a node $i\not\in V$
\STATE Fire node $i$
\ENDFOR
\ENDFOR
\RETURN $o$
\end{algorithmic}
\end{algorithm}

\subsubsection{Lack of Joint Distribution in Inference}
An ordinary learning machine constructs a joint distribution $\pi(X)$ as the estimation of the true distribution $p^*(X)$ in the learning phase and use $\pi(U|v)$ as the estimation of the posterior distribution $p^*(U|v)$ in the inference phase.
However, this approach often becomes intractable as the number of variables increases.

In contrast, the dependency network constructs $\pi_v(U)$---the stationary distribution of clamped-pseudo-Gibbs sampling--- every time an inference condition $V=v$ is given.
Therefore, any joint distribution $p(X)$ such that satisfies $\forall v,p(U|v)=\pi_v(U)$ does not exist.
Inference by pseudo-Gibbs sampling involves a trade-off.
Computation tractability is achieved; however, the existence of a joint distribution consistent for all inference queries is not achieved.
However, in many practical applications, users want to estimate the distribution $p^*(U|v)$ for each condition $V=v$, and the lack of joint distribution consistent for all inference queries does not become a significant disadvantage.

\section{Information Geometry of Pseudo-Gibbs Sampling}\label{sec:information geometry}
In Section \ref{sec:dependency}, we reviewed dependency networks and showed that pseudo-Gibbs sampling could be used for a tool to synthesize a joint distribution from CPTs.
In this section, we examine pseudo-Gibbs sampling from the perspective of information geometry.
As a result, we can interpret pseudo-Gibbs sampling as iterative m-projection onto certain manifolds.
Given this interpretation, we were motivated to consider making the stationary distribution close to any desired distribution.

\begin{figure}[H]
\centering\includegraphics{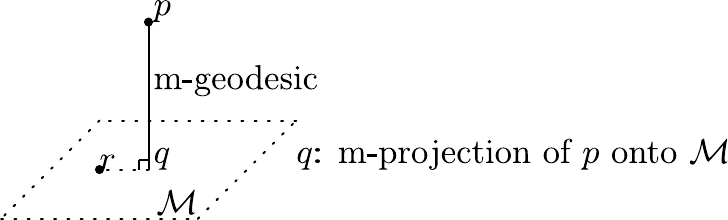}
\caption{m-projection onto e-flat manifold}\label{fig:m-projection}
\end{figure}

We here prepare some definitions and consequences in information geometry.
In information geometry, a distribution is represented by a point in a distribution space $\mathcal P:=\{p|p(X)\}$.
For two distributions $p_0,p_1\in\mathcal P$, the one-dimensional manifold
$$\{p_\lambda|p_\lambda(X)=(1-\lambda)p_0(X)+\lambda p_1(X),\lambda\in[0,1]\}$$
is called {\it m-geodesic} between $p_0,p_1$, and the one-dimensional manifold
\begin{align*}
\left\{p_\lambda\left|\ln p_\lambda(X)=(1-\lambda)\ln p_0(X)+\lambda\ln p_1(X)-\ln Z,\lambda\in[0,1]\right.\right\}\\
Z=\sum_x e^{(1-\lambda)\ln p_0(x)+\lambda\ln p_1(x)}
\end{align*}
is called {\it e-geodesic} between $p_0,p_1$ \citep[][Chapter 2]{amari2016information}.
For a manifold $\mathcal M$, if $\mathcal M$ includes any e-geodesic/m-geodesic between any two distributions $p_0,p_1\in\mathcal M$ then $\mathcal M$ is said to be {\it e-flat/m-flat}.
An m-geodesic $pq$ and an e-geodesic $qr$ are said to be {\it orthogonal} if and only if
$$\sum_x(p(x)-q(x))(\ln q(x)-\ln r(x))=0.$$
Let $KL(*\|*)$ denote the Kullback-Leibler divergence defined by
$$KL(p\|q):=\left<\ln\frac{p(X)}{q(X)}\right>_{p(X)}.$$
For a distribution $p\in\mathcal P$ and a manifold $\mathcal M$, {\it m-projection} of $p$ onto $\mathcal M$ is defined by
$$m_p(p,\mathcal M):=\arg\min_{q\in\mathcal M}KL(p\|q).$$
If $\mathcal M$ is e-flat then the minimizer $q$ is uniquely determined and m-geodesic $pq$ is orthogonal to any e-geodesic $qr(r\in\mathcal M)$ as illustrated in Figure \ref{fig:m-projection} \citep[][Chapter 2]{amari2016information}.\footnote{e-flat manifolds are denoted by dotted lines in figures in this paper.}

Now, we are ready to discuss the information geometry of actual/pseudo-Gibbs sampling.
We define the following manifold:
\begin{equation}\label{eq:E}
E(\theta_i):=\{p\in\mathcal P|p(X_i|X_{-i})=\theta_i(X_i|Y_i)\}
\end{equation}
and refer to it as {\it full-conditional-manifold} of node $i$.
Any distribution $p\in\mathcal P$ is expressed as $p(X_{-i})p(X_i|X_{-i})$, and the manifold in Eq.\eqref{eq:E} determines the second factor that is the full-conditional distribution.
Then, the following theorem holds.
\begin{theorem}\label{thm:e-flat}
Any full-conditional-manifold is e-flat and m-flat.
\end{theorem}
The proof is given in Appendix.

Let $X^0,X^1...$ be an output sequence of pseudo-Gibbs sampling.
Here, we consider pseudo-Gibbs sampling as movements of $p^t(=p(X^t))$ in $\mathcal P$.

Suppose that node $i$ fires at a certain time $t$.
Then, $p^t$ moves to the following distribution:
\begin{equation}\label{eq:move to}
p^{t+1}(X)=p^t(X_{-i})\theta_i(X_i|Y_i),
\end{equation}
that is, the full-conditional-distribution $p^t(X_i|X_{-i})$ is replaced by $\theta_i(X_i|Y_i)$.
As the following theorem shows, this $p^{t+1}$ is the m-projection of $p^t$ onto $E(\theta_i)$.
\begin{theorem}\label{thm:m-projection}
The m-projection of $p\in\mathcal P$ onto $E(\theta_i)$ is obtained by:
\begin{equation}\label{eq:m_p(p,E(theta_i))}
m_p(p,E(\theta_i))=p(X_{-i})\theta_i(X_i|Y_i).
\end{equation}
Therefore, firing node $i$ moves the distribution $p$ into the m-projection of $p$ onto $E(\theta_i)$.
\end{theorem}
The proof is given in Appendix.

\subsection{Ordered-(Actual/Pseudo)-Gibbs Sampling}
As described in Section \ref{sec:pseudo}, in pseudo-Gibbs sampling, there are two options to select a node to fire.
The theoretical analysis for ordered-pseudo-Gibbs sampling is complicated because ordered-pseudo-Gibbs sampling is an inhomogeneous Markov chain.
However, ordered-pseudo-Gibbs sampling behavior is easily understood when represented graphically.
\begin{figure}[H]
\centering\includegraphics{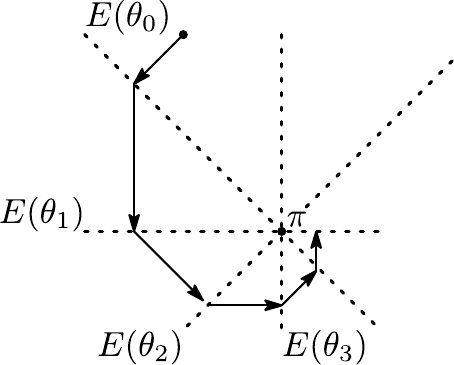}
\caption{Movement of $p^t$ in ordered-actual-Gibbs sampling}\label{fig:real-gs}
\end{figure}
\begin{figure}[H]
\centering\includegraphics{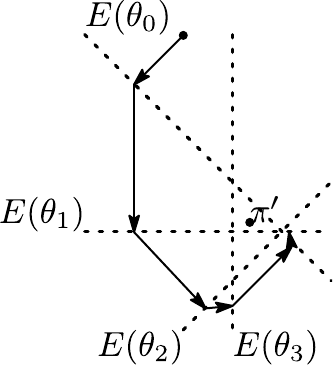}
\caption{Movement of $p^t$ in ordered-pseudo-Gibbs sampling}\label{fig:pseudo-gs}
\end{figure}
Figure \ref{fig:real-gs} illustrates the movements of $p^t$ in ordered-actual-Gibbs sampling.
In this case, the all full-conditional-manifolds intersect at a unique point $\pi$ because $\theta_i(X_i|Y_i)=\pi(X_i|X_{-i})$ for all $i$.
At first, node 0 fires, then the manifold $E(\theta_0)$ attracts $p^0$, and $p^0$ vertically falls down onto $E(\theta_0)$.
Next, node 1 fires, and $p^1$ falls down onto $E(\theta_1)$, and so on.
Given the depiction of the movement of $p^t$ in ordered-actual-Gibbs sampling shown in Figure \ref{fig:real-gs}, we can understand the convergence $p^t\to\pi$ intuitively.

On the other hand, Figure \ref{fig:pseudo-gs} illustrates the case of ordered-pseudo-Gibbs sampling.
In this case, the full-conditional-manifolds do not have a common intersection.
However, if we consider a situation where every full-conditional-manifold is close to a certain point $\pi'$, as in Figure \ref{fig:pseudo-gs}, $p^t$ is trapped in a small area around $\pi'$ after sufficient times of transitions.
To be more rigorous, $p^t$ moves along the cyclic orbit $\pi_0\to\pi_1\to...\to\pi_{n-1}\to\pi_0$, where $\pi_i$ is denoted in Eq.\eqref{eq:ordered}.
As shown in Eq.\eqref{eq:ordered}, $\pi$ is the centroid of $\{\pi_i\}(i=0,...,n-1)$.
Figure \ref{fig:pseudo-gs} gives us an intuition that $\pi$ is close to $\pi'$.

\subsection{Random-Pseudo-Gibbs Sampling}
Comparing with ordered-pseudo-Gibbs sampling, we can perform more formal analysis for random-pseudo-Gibbs sampling because it is a homogeneous Markov chain and has a stationary distribution.
This formal analysis supports Hypothesis 1 ``If every CPT $\theta_i(X_i|Y_i)$ is a good estimation of $p^*(X_i|X_{-i})$ then the stationary distribution of pseudo-Gibbs sampling becomes a good estimation of $p^*(X)$."

Under the condition that node $i$ fires at time $t$, $p^t$ moves to $p^{t+1}(X|i)=m_p(p^t,E(\theta_i))$ as shown in Theorem \ref{thm:m-projection}.
Therefore, under the condition that $i$ is unknown,
$$p^{t+1}(X)=\sum_i c_i p^{t+1}(X|i)=\sum_i c_i m_p(p^t,E(\theta_i)),$$
that is, a single step of random-pseudo-Gibbs sampling moves $p^t$ to $\sum_i c_i m_p(p^t,E(\theta_i))$.
Here, since $\pi$ is the stationary distribution,
\begin{equation}\label{eq:pi dont move}
\pi(X)=\sum_i c_im_p(\pi,E(\theta_i))(X)=\sum_i c_i\pi(X_{-i})\theta_i(X_i|Y_i)
\end{equation}
holds.

We here define KL-divergence between a distribution $p$ and a manifold $\mathcal M$ as
$$KL(p\|\mathcal M):=\min_{q\in\mathcal M}KL(p\|q)=KL(p\|m_p(p,\mathcal M))$$
and introduce the following pseudo-distance:
\begin{align}\label{eq:FC}
FC(p\|q)&:=\sum_i c_i KL(p\|E_i(q)),\quad E_i(q):=\{p\in\mathcal P|p(X_i|X_{-i})=q(X_i|X_{-i})\}\nonumber\\
&=\sum_i c_i\left<\ln\frac{p(X_i|X_{-i})}{q(X_i|X_{-i})}\right>_{p(X)}.
\end{align}
We refer to this pseudo-distance as {\it full-conditional-divergence}.
\begin{figure}[H]
\centering\includegraphics{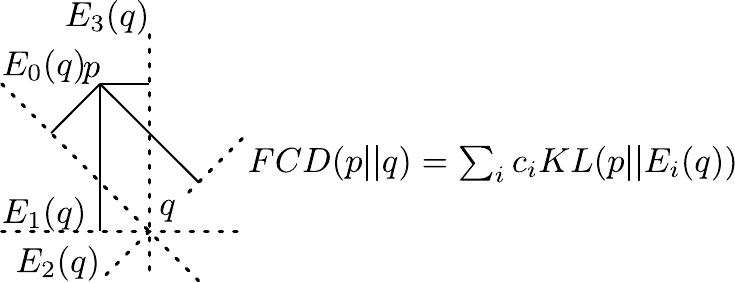}
\caption{$FC(p\|q)$}\label{fig:fcd}
\end{figure}
Figure \ref{fig:fcd} illustrates the geometrical interpretation of full-conditional-divergence, which is the average of KL-divergence between $p$ and the manifolds $E_i(q)$.
$FC(p||q)$ takes a finite value if it satisfies the condition $\forall x, p(x)>0\Rightarrow q(x)>0$.

The relationship between full-conditional-divergence and Besag's {\it pseudo-likelihood} \citep[][Chapter 20]{besag1975statistical,koller2009probabilistic} resembles the relationship between KL-divergence and likelihood.
We can rewrite Eq.\eqref{eq:FC} as
\begin{equation}\label{eq:FC rewrite}
FC(p\|q)=\left<\sum_ic_i\ln p(X_i|X_{-i})\right>_{p(X)}
-\underbrace{\left<\sum_ic_i\ln q(X_i|X_{-i})\right>_{p(X)}}
_{\text{pseudo-log-likelihood}}
\end{equation}
and the KL-divergence is given as:
\begin{equation}\label{eq:KL}
KL(p\|q)=\left<\ln p(X)\right>_{p(X)}
-\underbrace{\left<\ln q(X)\right>_{p(X)}}_{\text{log-likelihood}}.
\end{equation}
In Eqs.\eqref{eq:FC rewrite} and \eqref{eq:KL}, the first terms cancel the second terms when $p=q$.

Pseudo-(log-)likelihood was introduced by Besag as a convenient alternative to  (log-)likelihood to avoid the expensive computation of the partition function \citep{besag1975statistical}.
Note that minimizing KL-divergence is equivalent to maximizing likelihood; thus, minimizing full conditional divergence is equivalent to maximizing pseudo-likelihood.

Furthermore, KL-divergence and full-conditional-divergence are special cases of Bregman divergence \citep[][Chapter 2]{Censor97}.
Here, let $\Lambda$ be a convex subset of $\mathbb R^k$.
For a strictly convex differentiable function $f:\Lambda\to\mathbb R$, which is called the {\it Bregman function}, the Bregman divergence is defined as:
$$B_f(p\|q):=f(p)-f(q)-\nabla f(q)\cdot(p-q),$$
where $\nabla f(q)$ denotes the gradient vector of $f$ at $q$, and $\cdot$ denotes the Euclidean inner product.
We can treat a distribution $p\in\mathcal P$ as a vector in $\mathbb R^{|X|}$ whose $x$-th component is $p(x)$.
The KL-divergence is a Bregman divergence whose Bregman function is the negative entropy:
$$f(p)=-H_p(X)=\left<\ln p(X)\right>_{p(X)},$$
and the full-conditional-divergence is a Bregman divergence whose Bregman function is the average of negative conditional entropies:
$$f(p)=-\sum_i c_i H_p(X_i|X_{-i})=\sum_i c_i \left<\ln p(X_i|X_{-i})\right>_{p(X)}.$$

A noticeable feature of full-conditional-divergence is that it depends on the dependency among variables.
KL-divergence treats variables as a single concatenated variable; therefore, it does not reflect any dependency among variables.

The following theorem that indicates where the stationary distribution of random-pseudo-Gibbs sampling exists.
\begin{theorem}\label{thm:FC}
Let $p\in\mathcal P$ be an arbitrary distribution and $\pi$ be the stationary distribution of random-pseudo-Gibbs sampling using CPTs $\{\theta_i\}$.
Then, the following inequality holds.
$$FC(p\|\pi)\le\sum_ic_iKL(p\|E(\theta_i))$$
\end{theorem}
The proof is given in Appendix.
This theorem implies that if every full-conditional-manifold is close to a distribution $p$, then the stationary distribution $\pi$ is also close to $p$.
Thus, Hypothesis 1 is justified.

In the following, we focus on only random-pseudo-Gibbs sampling to simplify the theoretical analysis.

\section{Learning Algorithm Derived by Information Geometry}\label{sec:learning}
Assume a dependency network user seeks to approximate a given target distribution $p\in\mathcal P$ by the stationary distribution of pseudo-Gibbs sampling.
According to Theorem \ref{thm:FC}, the objective is to locate every full-conditional-manifold close to $p$, that is, the goal is to minimize $KL(p\|E(\theta_i))$ for each $i$.
This objective yields a structure/parameter learning algorithm.

Ideally, the target distribution should be the true distribution $p^*$; however, it is unknown.
Therefore, we employ $\tilde p^*$---empirical distribution of training data---as the target distribution and then consider certain regularization techniques to avoid overfitting to $\tilde p^*$.
Then, learning becomes a task to find inputs $Y_i$ and parameters $\theta_i(X_i|Y_i)$ that minimizes $KL(\tilde p^*\|E(\theta_i))+R_i$, where $R_i$ denotes a regularization term.
The minimization problem to be solved becomes as follows:
$$\min_{Y_i\subseteq X_{-i}}\min_{\theta_i}KL(\tilde p^*\|E(\theta_i))+R_i.$$

Here, we can divide this minimization problem into the following two stages of minimization.
\begin{description}
\item[Parameter learning.]Given $Y_i$, find $\hat\theta_i$ such that
\begin{equation}\label{eq:parameter learning}
\hat\theta_i=\arg\min_{\theta_i}KL(\tilde p^*\|E(\theta_i))+R_i.
\end{equation}
\item[Structure learning.]Find $\hat Y_i$ such that
\begin{equation}\label{eq:structure learning}
\hat Y_i=\arg\min_{Y_i\subseteq X_{-i}}KL(\tilde p^*\|E(\hat\theta_i))+R_i.
\end{equation}
\end{description}
We firstly determine inputs $Y_i$ by structure learning and then determine CPT by parameter learning.
For the convenience of explanation, we firstly discuss parameter learning and then discuss structure learning in this section.

\subsection{Parameter Learning}\label{sec:Parameter Learning}
We here prepare some quantities used in information theory.
For a distribution $p(X)$, the {\it entropy} $H_p(X)$ is defined by
$$H_p(X)=-\left<\ln p(X)\right>_{p(X)},$$
and for a distribution $p(XY)$, the {\it conditional entropy} $H_p(X|Y)$ is defined by
$$H_p(X|Y)=H_p(XY)-H_p(Y)=-\left<\ln p(X|Y)\right>_{p(XY)}.$$
Both quantities are always non-negative.

In Eq.\eqref{eq:parameter learning}, we use a regularization term $R_i$ that does not depend on the parameters; therefore, we can ignore $R_i$ in parameter learning.
Here, the following theorem holds.
\begin{theorem}\label{thm:KL(p||E)}
Let $\hat\theta_i(X_i|Y_i)$ be the CPT that minimizes $KL(p\|E(\theta_i(X_i|Y_i)))$ for given inputs $Y_i$.
Then, the following equations hold.
\begin{align}
&\hat\theta_i(X_i|Y_i)=p(X_i|Y_i)\label{eq:hat theta_i}\\
&KL(p\|E(\hat\theta_i))=H_p(X_i|Y_i)-H_p(X_i|X_{-i})\label{eq:KL(p||E)}.
\end{align}
\end{theorem}
The proof is given in Appendix.

Equation \eqref{eq:hat theta_i} directly provides a parameter learning algorithm.
Since $\tilde p^*$ is the empirical distribution of training data,
\begin{equation}\label{eq:counting data}
\hat\theta_i(x_i|y_i)=\frac{N_{x_iy_i}}{N_{y_i}},\quad
N_{y_i}=\sum_{x_i}N_{x_iy_i},
\end{equation}
where $N_{x_iy_i}$ is the number of occurrence of $X_i=x_i\wedge Y_i=y_i$ in the training data.
Then, the parameter learning algorithm becomes as follows:
\begin{algorithm}[H]
\caption{Parameter learning of node $i$}\label{algo:parameter learning}
$d$: training dataset\\
$Y_i$: inputs of node $i$\\
$\theta_i(X_i|Y_i)$: CPT
\begin{algorithmic}
\FORALL{$x_i,y_i$}
\STATE $N_{x_iy_i}\leftarrow\text{number of occurrences $X_i=x_i\wedge Y_i=y_i$ in $d$}$
\ENDFOR
\FORALL{$y_i$}
\STATE $N_{y_i}\leftarrow\sum_{x_i}N_{x_iy_i}$
\FORALL{$x_i$}
\STATE $\theta_i(x_i|y_i)\leftarrow N_{x_iy_i}/N_{y_i}$
\ENDFOR
\ENDFOR
\RETURN $\theta_i$
\end{algorithmic}
\end{algorithm}

\subsubsection{A Trick to Keep CPT Positive}
If $p^*$ is known, the objective function to be minimized in learning $p^*(X_i|X_{-i})$ is
$$KL(p^*\|E(\theta_i))=\left<\ln\frac{p^*(X)}{\theta_i(X_i|Y_i)}\right>_{p^*(X)}.$$
However, if CPT contains entries such that $\theta_i(x_i|y_i)=0$ then $KL(p^*\|E(\theta_i))$ diverges to infinity.
A simple trick to avoid this phenomenon is adding one to $N_{x_iy_i}$ such that $N_{x_iy_i}=0$.
Then, the parameter learning algorithm is modified as follows:
\begin{algorithm}[H]
\caption{Parameter learning of node $i$ (ensuring CPT positive)}\label{algo:parameter learning ver.2}
$d$: training dataset\\
$Y_i$: inputs of node $i$\\
$\theta_i$: CPT
\begin{algorithmic}[1]
\FORALL{$x_i,y_i$}
\STATE $N_{x_iy_i}\leftarrow\text{number of occurrences $X_i=x_i\wedge Y_i=y_i$ in $d$}$
\ENDFOR
\FORALL{$x_i,y_i$}\label{line:trick start}
\IF{$N_{x_iy_i}=0$}
\STATE $N_{x_iy_I}\leftarrow1$
\ENDIF
\ENDFOR\label{line:trick end}
\FORALL{$y_i$}
\STATE $N_{y_i}\leftarrow\sum_{x_i}N_{x_iy_i}$
\FORALL{$x_i$}
\STATE $\theta_i(x_i|y_i)\leftarrow N_{x_iy_i}/N_{y_i}$
\ENDFOR
\ENDFOR
\RETURN $\theta_i$
\end{algorithmic}
\end{algorithm}
The for-loop between line \ref{line:trick start} and line \ref{line:trick end} is the trick to ensure $N_{x_iy_i}>0$.

This trick provides another benefit.
It guarantees the ergodicity of pseudo-Gibbs sampling because, for any two values $x^0,x^1$, pseudo-Gibbs sampling is able to move from the state $X=x^0$ to the state $X=x^1$ within $n$ transitions.

\subsection{Structure Learning}\label{sec:Structure Learning)}
As shown in Eq.\eqref{eq:structure learning}, structure learning is the task to find $\hat Y_i\subseteq X_{-i}$ such that
\begin{align*}
\hat Y_i&=\arg\min_{Y_i\subseteq X_{-i}}KL(\tilde p^*\|E(\hat\theta_i))+R_i\\
&=\arg\min_{Y_i\subseteq X_{-i}}H_{\tilde p^*}(X_i|Y_i)-H_{\tilde p^*}(X_i|X_{-i})+R_i\quad\text{(by Eq.\eqref{eq:KL(p||E)})}\\
&=\arg\min_{Y_i\subseteq X_{-i}}H_{\tilde p^*}(X_i|Y_i)+R_i.
\end{align*}
Therefore, the cost function to be minimized in structure learning becomes as follows:
\begin{equation}\label{eq:scost_i}
scost_i(Y_i)=H_{\tilde p^*}(X_i|Y_i)+R_i.
\end{equation}
Here, $H_{\tilde p^*}(X_i|Y_i)$ is obtained by
$$H_{\tilde p^*}(X_i|Y_i)=-\sum_{x_iy_i}\frac{N_{x_iy_i}}N\ln\frac{N_{x_iy_i}}{N_{y_i}},
\quad N_{y_i}=\sum_{x_i}N_{x_iy_i}.$$
Various information criteria such as AIC \citep{akaike1974new} and MDL \citep{rissanen1989stochastic} are applicable for the regularization term:
\begin{align}
R_i&=\frac{k_i}N\quad\text{(AIC)}\label{eq:AIC}\\
R_i&=\frac{k_i}{2N}\ln N\quad\text{(MDL)},\label{eq:MDL}
\end{align}
where $k_i=(|X_i|-1)|Y_i|$ is the degrees of freedom that CPT $\theta_i(X_i|Y_i)$ has, and $|X_i|,|Y_i|$ are the number of values $X_i,Y_i$ possibly takes.

It is practical to use a greedy search to minimize $scost_i$.
The following algorithm is an example of greedy search for structure learning.
\begin{algorithm}[H]
\caption{Structure learning of node $i$}\label{algo:structure learning}
$Y_i$: inputs\\
$scost_i$: structure cost function (Eq.\eqref{eq:scost_i})
\begin{algorithmic}[1]
\STATE $Y_i\leftarrow\emptyset$
\LOOP
\STATE Evaluate $scost_i(Y'_i)$ for all possible candidates $Y'_i$, and find \label{line:candidate}
$Y^{min}_i=\arg\min_{Y'_i}scost_i(Y'_i)$
\IF{$scost_i(Y^{min})\ge scost_i(Y_i)$}\RETURN $Y_i$
\ELSE
\STATE $Y_i\leftarrow Y^{min}$
\ENDIF
\ENDLOOP
\end{algorithmic}
\end{algorithm}
``Candidate" in line \ref{line:candidate} is a set of input variables obtained by adding a single node into $Y_i$ or removing a single node from $Y_i$.

Other than this greedy algorithm, tabu search \citep[][]{glover1993user} is also practical to search $\hat Y_i$.

\subsection{Convergence to True Distribution}
A significant interest is determining whether stationary distribution $\pi$ converges to the true distribution $p^*$ at the limit of $N\to\infty$, where $N$ is the number of samples in the training data.

Here, let $d^\infty=\{x^0,x^1,...\}$ be a set of infinite training data, $d^N$ be the first $N$ samples of $d^\infty$ and $\tilde p^*(X)$ be the empirical distribution of $d^N$.
Given $d^N$, the structure/parameter learning algorithm of the node $i$ determines its inputs $Y_i$ and the parameters $\theta_i$; therefore $Y_i$ and $\theta_i$ are the functions of $N$.
Here, we consider the regularization term $R_i$ as a function $R_i(Y_i,N)$.
For simplicity, we assume that Algorithm \ref{algo:structure learning} actually finds the minimizer of Eq.\eqref{eq:scost_i} here.
The following theorem demonstrates that the full-conditional-manifolds approach $\tilde p^*$ as $N$ increases.
\begin{theorem}\label{thm:convergence of KL(bar p*||E)}
If $\lim_{N\to\infty}R_i(X_{-i},N)=0$ then
\begin{equation}\label{eq:convergence KL(p||E)}
\lim_{N\to\infty}KL(\tilde p^*\|E(\hat\theta_i))=0.
\end{equation}
\end{theorem}
The proof is given in Appendix.
It is clear that the regularization terms in Eqs.\eqref{eq:AIC} and \eqref{eq:MDL} satisfy the condition $\lim_{N\to\infty}R_i(X_{-i},N)=0$.

Note that the greedy search algorithm, such as Algorithm \ref{algo:structure learning}, may not find the global minimum of $scost_i$.
However, we can guarantee the convergence in Eq.\eqref{eq:convergence KL(p||E)} by employing a trick to such cases.
If we add the following step as the last step of any search algorithm then the convergence is guaranteed.
\begin{description}
\item[Last step)] If $scost_i(Y_i,N)>scost_i(X_{-i},N)$ then $Y_i\leftarrow X_{-i}$ and output $Y_i$.
\end{description}
However, this trick has less practical meaning because $scost_i(X_{-i},N)\gg scost_i(Y_i,N)$ in most practical cases.

According to Theorem \ref{thm:FC} and Theorem \ref{thm:convergence of KL(bar p*||E)}, it is clear that $\lim_{N\to\infty}FC(\tilde p^*\|\pi)=0$, and this result directly deduces the following theorem.
\begin{theorem}\label{thm:convergence to p^*}
If $\lim_{N\to\infty}R_i(X_{-i},N)=0$ and $\lim_{N\to\infty}\tilde p^*=p^*$ then
$$\lim_{N\to\infty}\pi=p^*.$$
\end{theorem}

\section{Inference by Pseudo-Gibbs Sampling}
According to Theorem \ref{thm:FC}, if $KL(p\|E(\theta_i))$ is small for each $i$, then the stationary distribution of pseudo-Gibbs sampling becomes a good estimation of $p(X)$.
As described in Section \ref{sec:inference}, clamped-pseudo-Gibbs sampling is equivalent to the free-pseudo-Gibbs sampling with CPTs ${\theta_i(X_i|Z_iv)}$ (see Section \ref{sec:inference} for the meaning of the symbols).
Then if $KL(p(\tilde V|v)\|\theta_i(X_i|Z_iv))$ is small for each $i$, the stationary distribution of the clamped-pseudo-Gibbs sampling becomes a good estimation of $p(\tilde V|v)$.

Consider the following equation:
\begin{align*}
KL(p(X)\|E(\theta_i(X_i|X_{-i})))&=\left<\ln\frac{p(X_i|X_{-i})}{\theta_i(X_i|X_{-i})}\right>_{p(X)}\\
&=\sum_{x_iz_iv} p(x_iz_iv)\ln\frac{p(x_i|z_iv)}{\theta_i(x_i|z_iv)}\\
&=\sum_v p(v)\sum_{x_iz_i}p(x_iz_i|v)\ln\frac{p(x_i|z_iv)}{\theta_i(x_i|z_iv)}\\
&=\sum_v p(v)\left<\ln\frac{p(X_i|Z_iv)}{\theta_i(X_i|Z_iv)}\right>_{p(\tilde V|v)}\\
&=\sum_v p(v)KL(p(\tilde V|v)\|E(\theta_i(X_i|Z_iv)).
\end{align*}
This equation means that $KL(p\|E(\theta_i))$ is the average of $KL(p(\tilde V|v)\|E(\theta_i(X_i|Z_iv))$.
In other words, if $KL(p\|E(\theta_i))$ is small then $KL(p(\tilde V|v)\|E(\theta_i(X_i|Z_iv)))$ is small on average.
Note that $KL(p(\tilde V|v)\|E(\theta_i(X_i|Z_iv))$ may become large if $p(v)$ is small, which implies that the inference for rare conditions becomes difficult.

\section{Generalized Input for CPT}
In the sections so far, we let the inputs $Y_i$ be certain variables in $X_{-i}$.
We refer to this type of inputs as the {\it direct input}.
Here, we notice it is unnecessary to let $Y_i$ be variables in $X_{-i}$ because $Y_i$ is just a (concatenated) variable to be referred by CPT.
Even if we consider $Y_i$ is a variable determined by $Y_i=f_i(X_{-i})$, where $f_i$ is a function of $X_{-i}$, the most of discussion made in the sections so far remains valid.
Let us refer to this type of input as the {\it generalized input}.
The direct input is a special case of the generalized input where $f_i$ is a function to select variables in $X_{-i}$.

In the case we use the generalized input, the role of structure learning is to find a function $f_i$ that minimizes the structure cost in Eq.\eqref{eq:scost_i}.
Note that $|f_i|$, which is the size of $f_i$'s range, is suppressed to suppress the CPT's degrees of freedom and the regularization term.

Providing a function $f_i$ is equivalent to clustering all possible values $x_{-i}$ into $|f_i|$ (size of $f_i$'s range) groups.
A practical method to find $f_i$ in structure learning is using a decision tree learning algorithm \citep{heckerman2000dependency}.
In this case, a leaf of the decision tree corresponds to a group that shares a value of $f_i$.

\section{Experiments}\label{sec:experiments}
In this section, we demonstrate a dependency network's performance by comparing it to a Bayesian network.

\subsection{Compared Learning Machines}
We compared two learning machines: a dependency network (DN) and a Bayesian network (BN).

\subsubsection{Dependency Network DN}\label{sec:DN}
We used Algorithm \ref{algo:structure learning} for structure learning and Algorithm \ref{algo:parameter learning ver.2} for parameter learning.
In structure learning, we used MDL in Eq.\eqref{eq:MDL} for the regularization term.
For inference, we used Algorithm \ref{algo:clamped with burn-in and thinning} and used the setting $b=k=n$(number of nodes) for burn-in and thinning.

\subsubsection{Bayesian Network BN}
Similarly to the learning discussed in Section \ref{sec:learning}, learning in BN comes down to the following minimization problems.
\begin{description}
\item[Parameter learning.]Given graphical structure $G$, find $\hat\pi$ such that
$$\hat\pi=\arg\min_\pi KL(\tilde p^*\|\pi)+R.$$
\item[Structure learning.]Find $\hat G$ such that
$$\hat G=\arg\min_G KL(\tilde p^*|\hat\pi)+R.$$
\end{description}
We here skip the derivation and only show the following consequences.
\begin{equation}\label{eq:BN parameter learning}
\hat\pi(X_i|Y_i)=\tilde p^*(X_i|Y_i)
\end{equation}
$$KL(\tilde p^*\|\hat\pi)=-H_{\tilde p^*}(X)+\sum_i H_{\tilde p^*}(X_i|Y_i)$$
\begin{equation}\label{eq:BN scost}
scost(G)=\sum_i H_{\tilde p^*}(X_i|Y_i)+R
\end{equation}
In the experiments, we used the following MDL based regularization term.
$$R=\sum_i\frac{k_i}{2N}\ln N,\quad k_i=(|X_i|-1)|Y_i|,$$
where $k_i$ is the degrees of freedom that CPT of node $i$ has (see Eq.\eqref{eq:MDL}).

We used the following algorithm for structure and parameter learning.
\begin{algorithm}[H]
\caption{Structure and parameter learning in BN}\label{algo:BN learning}
$G$: graph of BN\\
$N$: number of samples in training data\\
$scost$: structure cost function (Eq.\eqref{eq:BN scost})
\begin{algorithmic}[1]
\STATE $G\leftarrow\emptyset$
\LOOP
\STATE Evaluate $scost(G')$ for all possible candidates $G'$, and find\\
$G^{min}=\arg\min_{G'}scost(G')$ \label{line:BN candidate}
\IF{$scost(G^{min})\ge scost(G)$}
\STATE Determine CPT by Eq.\eqref{eq:BN parameter learning}
\RETURN $G$ and CPT
\ENDIF
\STATE $G\leftarrow G^{min}$
\ENDLOOP
\end{algorithmic}
\end{algorithm}
``Candidate" in line \ref{line:BN candidate} is a graph obtained by one of the following modifications to $G$.
\begin{itemize}
\item Adding a single edge to $G$ unless the candidate has any cycles.
\item Removing a single edge from $G$.
\item Reversing the direction of a single edge in $G$ unless the candidate has any cycles.
\end{itemize}

We drew output data by the following algorithm.
\begin{algorithm}[H]
\caption{Drawing output data from BN}\label{algo:BN output}
$G$: graph of BN\\
$H$: a graph\\
$o$: output data\\
$a_*$: sequence of node number\\
$n$: number of nodes\\
$N$: number of samples in output data
\begin{algorithmic}[1]
\STATE $H\leftarrow G$, $o\leftarrow\emptyset$
\FOR{$i=0$ to $n-1$}\label{line:topological sort start}
\STATE Select a node $j$ such that has no parents in $H$
\STATE $a_i\leftarrow j$;
\STATE Remove node $j$ and its outgoing edges from $H$
\ENDFOR\label{line:topological sort end}
\FOR{$t=0$ to $N-1$}
\FOR{$i=0$ to $n-1$}
\STATE Fire node $a_i$ \label{line:fire}
\ENDFOR
\STATE Append $X$ into $o$
\ENDFOR
\RETURN $o$
\end{algorithmic}
\end{algorithm}
The for-loop between line \ref{line:topological sort start} and line \ref{line:topological sort end} is a {\it topological sorting} \citep[section 2.2.3]{knuth1997art}.

The output data drawn by this algorithm are i.i.d. data; thus, neither burn-in nor thinning techniques are needed.

\subsection{Training Datasets}\label{sec:datasets}
We used the following four training datasets for the experiments.
Each dataset is an i.i.d. dataset drawn from a known true distribution $p^*$.

\subsubsection{BN20-37S, BN20-37L}
\begin{figure}[H] 
\centering\includegraphics{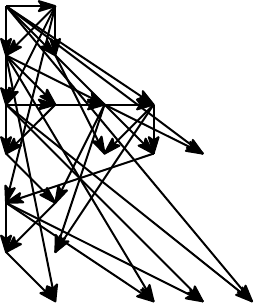}
\caption{BN20-37}\label{fig:BN20-37}
\end{figure}
BN20-37S and BN20-37L are datasets drawn from a Bayesian network with 20 nodes and 37 edges (Figure \ref{fig:BN20-37}).
Every $X_i$ takes value 0 or 1.
BN20-37S includes 1000 samples, and BN20-37L has 100000 samples.

\subsubsection{Ising5x5S, Ising5x5L}
\begin{figure}[H]
\centering\includegraphics{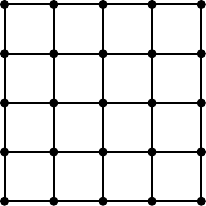}
\caption{Ising5x5}\label{fig:Ising5x5}
\end{figure}
Ising5x5S and Ising5x5L are datasets drawn from a Ising spin model that has $5\times5$ nodes (Figure \ref{fig:Ising5x5}).
Every $X_i$ takes the value 0 or 1.
Ising5x5S has 1000 samples, and Ising5x5L has 100000 samples.

\subsection{Platform for Experiments}
All experiments were performed on a laptop PC(CPU: Intel Core i7-6700K @4GHz; Memory: 64GB; OS: Windows 10 Pro).
All programs were written in and executed on Java 8.

\subsection{Results}
\begin{table}[h]
\centering
\caption{Performance comparison between DN and BN}\label{table:DN and BN}
\begin{tabular}{|c|c|r|c|c|c|} \hline
Dataset &$n$ &$N$ &$KL(\tilde p^*\|p^*)$ &$KL(\tilde\pi\|p^*)$(DN)&$KL(\tilde\pi\|p^*)$(BN)\\ \hline
					&		&				&			&1.43	&0.78\\
BN20-37S	&20	&1000		&4.14	&1.43	&0.78\\
					&		&				&			&1.43	&0.78\\ \hline
					&		&				&			&0.69	&0.69\\
BN20-37L	&20	&100000	&0.67	&0.69	&0.69\\
					&		&				&			&0,69	&0.70\\ \hline
					&		&				&			&1.37	&1.37\\
Ising5x5S	&25	&1000		&3.87	&1.36	&1,37\\
					&		&				&			&1.39	&1.37\\ \hline
					&		&				&			&1.17	&1.16\\
Ising5x5L	&25	&100000	&1.16	&1.18	&1.15\\
					&		&				&			&1.19	&1.16\\ \hline
\end{tabular}\label{table:KL}\\
$n$: number of variables.
$N$: number of samples.
$\tilde p^*$: empirical distribution of training data.
$p^*$: true distribution. $\tilde\pi$: empirical distribution of output data.
The unit for KL-divergence is ``nat".
\end{table}
Table \ref{table:DN and BN} demonstrates the performance comparison between DN and BN.
After learning, we drew 1000000 samples of output data respectively from DN and BN.
We measured the accuracy of learning by $KL(\tilde\pi\|p^*)$, where $\tilde\pi$ is the empirical distribution of output data, and $p^*$ is the true distribution of the training dataset.
The outputs depend on the random seeds used to draw outputs; therefore, we made three trials with different random seeds for each dataset.

For BN20-27S, BN has better performance than DN has.
One possible reason for this result is that BN20-37S is a dataset drawn from a  Bayesian network, so BN easily learns $p^*$ while DN can not imitate $p^*$ by a simple structure.
For the larger samples (BN20-37L), DN is allowed to take a more complex structure and learn $p^*$ correctly.

For the datasets BN20-37L, Ising5x5S, and Ising5x5L, $KL(\tilde\pi\|p^*)(DN)\approx KL(\tilde\pi\|p^*)(BN)$, that is, DN and BN have roughly the same performance.

For the datasets BN20-37L and Ising5x5L, both the training data and the output data have the same number of samples (100000); therefore, we can compare $KL(\tilde\pi\|p^*)$ and $KL(\tilde p^*\|p^*)$ directly.
The fact $KL(\tilde\pi\|p^*)\approx KL(\tilde p^*\|p^*)$ implies that the output data drawn from $\pi$ and the training data---i.i.d. data drawn from $p^*$---are indistinguishable; thus, the learning machine correctly learns $p^*$.

\begin{table}[h]
\centering
\caption{Learning behavior of node 0 in DN}\label{table:node 0}
\begin{tabular}{|c|l|c|c|c|c|} \hline
Dataset	&$Y_0$	&$H(X_0)$	&$H(X_0|Y_0)$	&$KL(\tilde p^*\|E(\theta_0))$ &$KL(p^*\|E(\theta_0))$\\ \hline
BN20-37S  &$X_1X_2X_4X_{19}$								&0.68	&0.41	&0.41	&0.17\\
BN20-37L  &$X_1X_2X_3X_4X_6X_7X_{11}X_{19}$	&0.68	&0.22	&0.09	&1.3E-3\\
Ising5x5S	&$X_1X_5$													&0.69	&0.44	&0.36	&2.9E-3\\
Ising5x5L &$X_1X_5$													&0.69	&0.44	&0.19	&1.4E-5\\ \hline
\end{tabular}\\
The unit for entropy and KL-divergence is ``nat".
\end{table}

Here, we examine more details of the learning behavior of DN.
Table \ref{table:node 0} focuses on node 0.
We explain the results for BN20-37S as an example here.
The learning algorithm starts with no inputs ($Y_0=\emptyset$), and $X_0$ has 0.68nat of entropy.
The structure learning algorithm takes $X_1X_2X_4X_{19}$ as the inputs, and the entropy of $X_0$ is reduced to 0.41nat.
For the full conditional manifold $E(\theta_0)$ constructed by the learning algorithm, $KL(\tilde p^*\|E(\theta_0))=0.41$nat and $KL(p^*\|E(\theta_0))=0.17$nat.
The fact $KL(\tilde p^*\|E(\theta_i))\gg KL(p^*\|E(\theta_i))$ implies that overfitting to $\tilde p^*$ is successfully avoided, and the learning algorithm has high generalization ability.
We confirmed that the inequality $KL(\tilde p^*\|E(\theta_i))\gg KL(p^*\|E(\theta_i)))$ holds for every node.
\begin{table}[h]
\centering
\caption{Average of $KL(\tilde p^*\|E(\theta_i))$ and $KL(p^*\|E(\theta_i))$}\label{table:average KL}
\begin{tabular}{|c|c|c|} \hline
Dataset &$\overline{KL(\tilde p^*\|E(\theta_0))}$ &$\overline{KL(p^*\|E(\theta_0))}$\\ \hline
BN20-37S	&0.47	&4.5E-2\\
BN20-37L	&0.17	&1.1E-3\\
Ising5x5S &0.28	&7.0E-3\\
Ising5x5L &0.16	&5.3E-5\\ \hline
\end{tabular}\\
The unit for entropy and KL-divergence is ``nat".
\end{table}

Table \ref{table:average KL} demonstrates the average of $KL(\tilde p^*\|E(\theta_i))$ and $KL(p^*\|E(\theta_i))$ defined by:
$$\overline{KL(\tilde p^*\|E(\theta_i))}=\sum_i c_i KL(\tilde p^*\|E(\theta_i)),\quad
\overline{KL(p^*\|E(\theta_i))}=\sum_i c_i KL(p^*\|E(\theta_i)).$$
As shown in the table, $\overline{KL(p^*\|E(\theta_i))}$ takes small values, and this fact implies that the stationary distribution $\pi$ is close to the true distribution $p^*$ because the inequality $FC(p^*\|\pi)\le\overline{KL(p^*\|E(\theta_i))}$ holds by Theorem \ref{thm:FC}.

\begin{table}[h]
\caption{Number of structure cost evaluations and CPU time for learning}\label{table:learning time}
\begin{center}
\begin{tabular}{|c|c|r|r|r|r|r|} \hline
Dataset		&$n$	&$N$		&\#eval(DN)	&\#eval(BN)	&time(DN)	&time(BN)\\ \hline
BN20-37S	&20		&1000		&1616				&16945 			&17				&73\\
BN20-37L	&20		&100000 &2129				&23797			&2466			&13127\\
Ising5x5S	&25		&1000		&2497				&32298			&25				&144\\
Ising5x5L	&25		&100000 &2545				&42365			&2419			&22676\\ \hline
\end{tabular}\label{table:time}\\
$n$: number of nodes. $N$: number of samples in training data.\\
\#eval: number of structure cost evaluations. time: CPU time to learn (ms).
\end{center}
\end{table}
Table \ref{table:learning time} demonstrates the number of structure cost evaluations and CPU time to learn the datasets.
Here, \#eval(DN) is the number of evaluations in Algorithm \ref{algo:structure learning} line \ref{line:candidate}, and \#eval(BN) is the number of evaluations in Algorithm \ref{algo:BN learning} line \ref{line:BN candidate}.
To measure CPU time, we made three trials for each dataset and showed the median in the table.

In the learning algorithms used for DN and BN, the structure cost evaluations dominate the computational costs.
The computational cost to evaluate one structure obeys $O(N)$; thus, the time to learn obeys $O(N\times\text{\#eval})$.
As shown in the table, DN learns the datasets with much smaller number of evaluations than BN and learns faster than BN.

\section{Conclusion and Future Work}
In dependency network learning, the network does not directly learn the true distribution $p^*(X)$ underlying training data.
Rather than learning the joint distribution $p^*(X)$, each node independently learns the full conditional distribution $p^*(X_i|X_{-i})$.
This property---independence of learning tasks---keeps the dependency network learning tractable even where the number of nodes is large.

Pseudo-Gibbs sampling gathers the CPTs owned by the nodes and constructs a joint distribution as its stationary distribution.
If every node learns $p^*(X_i|X_{-i})$ successfully, that is, every CPT is a good estimation of $p^*(X_i|X_{-i})$, then the stationary distribution of pseudo-Gibbs sampling becomes a good estimation of $p^*(X)$.
It is challenging to know the stationary distribution numerically, however, we can draw as many samples as we want from the stationary distribution using pseudo-Gibbs sampling; thus, we can estimate the probabilities and expectation of variables using Monte Carlo methods.

From the information geometry perspective, we consider a manifold $E(\theta_i)$ for each node determined by its CPT.
Then, pseudo-Gibbs sampling is interpreted as iterative m-projections onto these manifolds.
This interpretation tells us how we should make the stationary distribution of pseudo-Gibbs sampling close to the desired distribution and simultaneously provides the structure/parameter learning algorithms as optimization problems.

We introduced a pseudo-distance $FC(p\|q)$ that is a Bregman divergence whose Bregman function is Besag's pseudo-log-likelihood.
We built a hypothesis ``If every CPT is a good estimation of $p^*(X_i|X_{-i})$ then the stationary distribution of pseudo-Gibbs sampling is close to $p^*(X)$."
Using the pseudo-distance $FC$, we justified this hypothesis (Theorem \ref{thm:FC}).

We compared a dependency network (DN) and a Bayesian network (BN) experimentally.
Regarding the accuracy of the learned distribution, there was no significant difference between DN and BN.
The experiments also demonstrated that DN was much faster than BN in learning.

One possible future work is performance evaluation by real problems.
Suppose we perform a hold-out test for performance evaluation.
For real problems, the true distribution $p^*$ is unknown, and only $p^{test}$ that is an empirical distribution of data drawn from $p^*$ is known.
In the case of a dependency network, the learned distribution $\pi$ is also unknown, and only $\tilde\pi$, which is the empirical distribution of data drawn from $\pi$, is given.
Furthermore, both $p^{test}$ and $\tilde\pi$ are very ``sparse", that is, $p^{test}(x)=0$ and $\tilde\pi(x)=0$ for most values $x$.
Therefore, we have to compare two sparse distributions $p^{test}$ and $\tilde\pi$.
However, evaluating the closeness between such two sparse distribution is a challenging problem.

Another possible future work is introducing hidden variables into dependency networks.
We expect that introducing hidden variables will provide much higher performance into dependency networks.


\acks{This work was supported by JSPS KAKENHI 17H01793.} 


\appendix
\section*{Appendix}
\subsection*{Proof of Theorem \ref{thm:e-flat}}
Let $p_0,p_1$ be two distributions on the manifold $E(\theta_i)$, that is,
\begin{align*}
p_0(X)&=p_0(X_{-i})\theta_i(X_i|Y_i)\\
p_1(X)&=p_1(X_{-i})\theta_i(X_i|Y_i)
\end{align*}
and $p_\lambda$ be a distribution on the e-geodesic between $p_0$ and $p_1$ expressed by
\begin{align*}
\ln p_\lambda(X)&=(1-\lambda)\ln p_0(X)+\lambda p_1(X)-\ln Z\\
&=\ln(1-\lambda)p_0(X_{-i})+\lambda p_1(X_{-i})+\ln\theta_i(X_i|Y_i)-\ln Z.
\end{align*}
Then, we obtain:
$$p_\lambda(X)=\frac{\theta_i(X_i|Y_i)}Z e^{(1-\lambda)p_0(X_{-i})
+\lambda p_1(X_{-i})}$$
and
\begin{align*}
p_\lambda(X_{-i})&=\sum_{X_i}p_\lambda(X)\\
&=\frac1Z e^{(1-\lambda)p_0(X_{-i})+\lambda p_1(X_{-i})}\\
p_\lambda(X_i|X_{-i})&=\frac{p_\lambda(X)}{p_\lambda(X_{-i})}=\theta_i(X_i|Y_i).
\end{align*}
Therefore, $E(\theta_i(X_i|Y_i))$ is e-flat.

Similarly, let $q_\lambda$ be a distribution on the m-geodesic between $p_0$ and $p_1$ expressed by
\begin{align*}
q_\lambda(X)
&=(1-\lambda)p_0(X)+\lambda p_1(X)\\
&=((1-\lambda)p_0(X_{-i})+\lambda p_1(X_{-i}))\theta_i(X_i|Y_i).
\end{align*}
Then, we obtain:
$$q_\lambda(X_{-i})=(1-\lambda)p_0(X_{-i})+\lambda p_1(X_{-i})$$
and
$$q_\lambda(X_i|X_{-i})=\frac{q_\lambda(X)}{q_\lambda(X_{-i})}
=\theta_i(X_i|Y_i).$$
Therefore, $E(\theta_i)$ is m-flat.\qed

\subsection*{Proof of Theorem \ref{thm:m-projection}}
Let $q(X)=q(X_{-i})\theta_i(X_i|Y_i)$ be a distribution on $E(\theta_i)$.
Then, we obtain:
\begin{align*}
KL(p\|q)&=\left<\ln\frac{p(X)}{q(X)}\right>_{p(X)}\\
&=\left<\ln\frac{p(X_{-i})}{q(X_{-i})}\right>_{p(X)}
+\left<\ln\frac{p(X_i|X_{-i})}{\theta_i(X_i|Y_i)}\right>_{p(X)}\\
&=KL(p(X_{-i})\|q(X_{-i}))
+\left<\ln\frac{p(X_i|X_{-i})}{\theta_i(X_i|Y_i)}\right>_{p(X)}.
\end{align*}
It is clear that $KL(p\|q)$ is minimized when $q(X_{-i})=p(X_{-i})$; thus, we obtain:
$$m_p(p,E(\theta_i))=\arg\min_{q\in E(\theta_i)}KL(p\|q)=p(X_{-i})\theta_i(X_i|Y_i).$$
This result coincides with Eq.\eqref{eq:move to}.
\qed

\subsection*{Proof of Theorem \ref{thm:FC}}
Let $\pi_i$ denote the m-projection of $\pi$ onto $E(\theta_i)$, that is,
$\pi_i(X)=\pi(X_{-i})\theta_i(X_i|Y_i)$.
Then, we obtain:
\begin{align*}
\sum_i c_i KL(p\|E(\theta_i))-FC(p\|\pi)
&=\sum_i c_i\left<\ln\frac{p(X_i|X_{-i})}{\theta_i(X_i|Y_i)}\right>_{p(X)}
-\sum_i c_i\left<\ln\frac{p(X_i|X_{-i})}{\pi(X_i|X_{-i})}\right>_{p(X)}\\
&=\sum_i c_i\left<\ln\frac{\pi(X_i|X_{-i})}{\theta_i(X_i|Y_i)}\right>_{p(X)}\\
&=\sum_i c_i\left<\ln\frac{\pi(X_{-i})\pi(X_i|X_{-i})}{\pi(X_{-i})\theta_i(X_i|Y_i)}\right>_{p(X)}\\
&=\sum_i c_i\left<\ln\frac{\pi(X)}{\pi_i(X)}\right>_{p(X)}\\
&=\left<-\sum_i c_i\ln\frac{\pi_i(X)}{\pi(X)}\right>_{p(X)}\\
&\ge\left<-\ln\sum_i c_i\frac{\pi_i(X)}{\pi(X)}\right>_{p(X)}\quad
\text{$\because-\ln$ is a convex function}\\
&=\left<-\ln\frac{\pi(X)}{\pi(X)}\right>_{p(X)}\quad\text{by Eq.\eqref{eq:pi dont move}}\\
&=0.
\end{align*}
\qed

\subsection*{Proof of Theorem \ref{thm:KL(p||E)}}
\begin{align*}
KL(p\|E(\theta_i(X_i|Y_i)))&=KL(p\|m_p(p,E(\theta_i))\\
&=\left<\ln\frac{p(X_i|X_{-i})}{\theta_i(X_i|Y_i)}\right>_{p(X)}\\
&=\left<\ln\frac{p(X_i|Y_i)}{\theta_i(X_i|Y_i)}\right>_{p(X)}
+\left<\ln\frac{p(X_i|X_{-i})}{p(X_i|Y_i)}\right>_{p(X)}\\
&=\left<\ln\frac{p(X_i|Y_i)}{\theta_i(X_i|Y_i)}\right>_{p(X)}+H_p(X_i|Y_i)
-H_p(X_i|X_{-i})\\
&=\sum_{y_i}p(y_i)KL(p(X_i|y_i)\|\theta_i(X_i|y_i))+H_p(X_i|Y_i)-H_p(X_i|X_{-i}).
\end{align*}
It is clear that this equation is minimized when $\theta_i(X_i|Y_i)=p(X_i|Y_i)$, and then
$$KL(p\|E(\theta_i(X_i|Y_i)))=H_p(X_i|Y_i)-H_p(X_i|X_{-i}).$$
\qed

\subsection*{Proof of Theorem \ref{thm:convergence of KL(bar p*||E)}}
Since $\hat Y_i$ is the minimizer of the cost function $scost$ in Eq.\eqref{eq:scost_i}, $scost_i(\hat Y_i,N)\le scost_i(X_{-i},N)$, that is,
$$H_{\tilde p^*}(X_i|\hat Y_i)+R_i(\hat Y_i,N)\le H_{\tilde p^*}(X_i|X_{-i})+R_i(X_{-i},N).$$
Therefore, we obtain:
\begin{equation}\label{eq:bound KL}
\begin{split}
KL(\tilde p^*\|E(\hat\theta_i))&=H_{\tilde p^*}(X_i|\hat Y_i)-H_{\tilde p^*}(X_i|X_{-i})\quad\text{by Eq.\eqref{eq:KL(p||E)}}\\
&\le R_i(X_{-i},N)-R_i(\hat Y_i,N)\\
&\le R_i(X_{-i},N).
\end{split}
\end{equation}
Considering Eq.\eqref{eq:bound KL} at the limit of $N\to\infty$, we obtain:
$$0\le\lim_{N\to\infty}KL(\tilde p^*\|E(\hat\theta_i))\le\lim_{N\to\infty}R_i(X_{-i},N)=0.$$
\qed

\vskip 0.2in
\bibliography{myref}

\end{document}